\documentclass[11pt]{article}

\usepackage[utf8]{inputenc} % Input encoding
\usepackage[T1]{fontenc}    % Font encoding
\usepackage{graphicx}       % Including figures (though not used here)
\usepackage{amsmath}        % Math environments
\usepackage{amssymb}        % Math symbols
\usepackage{booktabs}       % Professional tables (\toprule, \midrule, \bottomrule)
\usepackage[numbers, sort&compress]{natbib} % Citation management (using numbers for arXiv, but natbib commands work)
\usepackage{hyperref}       % Clickable links and references
\usepackage{caption}        % Better control over captions
\usepackage{geometry}       % Adjust margins if needed (optional)
\usepackage{rotating}

\hypersetup{
    colorlinks=true,
    linkcolor=blue,
    filecolor=magenta,
    urlcolor=cyan,
    citecolor=blue, % Color for citations if using colorlinks
    pdftitle={Lexical Bundle Frequency as a Construct-Relevant Candidate Feature in Automated Scoring of L2 Academic Writing},
    pdfauthor={Burak Senel}, % Add author names here
    pdfsubject={Computational Linguistics, Language Assessment},
    pdfkeywords={Automated Scoring, Lexical Bundles, Natural Language Processing, Language Testing, TOEFL, L2 Writing},
}

\title{Lexical Bundle Frequency as a Construct-Relevant Candidate Feature in Automated Scoring of L2 Academic Writing}
\author{%
  Burak Senel \\
  \small Iowa State University \\
  \small \texttt{senel@iastate.edu}
}
\date{\today} % Or specify a date: \date{October 26, 2023}

\begin{document}

\maketitle

\begin{abstract}
Automated scoring (AS) systems are increasingly used for evaluating L2 writing, but require ongoing refinement for construct validity. While prior work suggested lexical bundles (LBs) – recurrent multi-word sequences satisfying certain frequency criteria – could inform assessment, their empirical integration into AS models needs further investigation. This study tested the impact of incorporating LB frequency features into an AS model for TOEFL independent writing tasks. Analyzing a sampled subcorpus (N=1,225 essays, 9 L1s) from the TOEFL11 corpus, scored by ETS-trained raters (Low, Medium, High), 3- to 9-word LBs were extracted, distinguishing prompt-specific from non-prompt types. A baseline Support Vector Machine (SVM) scoring model using established linguistic features (e.g., mechanics, cohesion, sophistication) was compared against an extended model including three aggregate LB frequency features (total prompt, total non-prompt, overall total). Results revealed significant, though generally small-effect, relationships between LB frequency (especially non-prompt bundles) and proficiency (p < .05). Mean frequencies suggested lower proficiency essays used more LBs overall. Critically, the LB-enhanced model improved agreement with human raters (Quadratic Cohen’s Kappa +2.05\%, overall Cohen's Kappa +5.63\%), with notable gains for low (+10.1\% exact agreement) and medium (+14.3\% Cohen’s Kappa) proficiency essays. These findings demonstrate that integrating aggregate LB frequency offers potential for developing more linguistically informed and accurate AS systems, particularly for differentiating developing L2 writers.
\end{abstract}

% --- SECTIONS ---

\section{Introduction}
\label{sec:intro}
The evaluation of second language (L2) writing proficiency is a critical component of language assessment, particularly in high-stakes contexts like university admissions or placement. The \emph{Test of English as a Foreign Language} (TOEFL) independent writing task, requiring timed essay production, is a widely used example. Traditionally scored by human raters, the logistical demands and costs associated with large-scale assessment have driven the adoption of automated scoring (AS) systems, such as ETS's e-rater \citep{Attali2006}.

AS systems leverage computational methods to analyze text features (e.g., grammar, vocabulary, organization) and predict scores aligned with human judgments. While offering efficiency and consistency benefits, a primary challenge for AS is ensuring construct validity – verifying that the system measures the intended writing abilities accurately and comprehensively \citep{Xi2010}. Relying solely on easily quantifiable features like essay length can lead to models that correlate with human scores but fail to capture deeper linguistic competence \citep{Attali2006}. Therefore, ongoing research focuses on identifying and integrating more sophisticated linguistic features that better represent the complexities of proficient writing.

One promising area involves the analysis of lexical bundles (LBs) – statistically frequent, recurrent sequences of three or more words (e.g., "on the other hand," "as a result of," "it is important to"). LBs are considered building blocks of fluent discourse and are prevalent in various registers, including academic writing \citep{Biber1999, Biber2004}. Crucially for assessment, research suggests that the frequency and type of LBs used by L2 writers correlate with their proficiency levels. Less proficient writers may overuse certain bundles, particularly those directly cued by the prompt, potentially reflecting reliance on less analyzed linguistic chunks \citep{Ellis2002, Staples2013, Vo2019}.

While previous studies have highlighted correlations between LB use and proficiency and suggested their potential for assessment \citep{Appel2016, Staples2013, Vo2019}, few have empirically investigated the impact of \emph{integrating} LB features directly into AS models, particularly within a diverse, high-stakes testing context like TOEFL. This study aims to fill this gap by:
1.  Examining the relationship between the frequency of various LB types (3-9 words, prompt-specific vs. non-prompt) and human-assigned proficiency scores in a large corpus of TOEFL essays.
2.  Evaluating the extent to which adding aggregate LB frequency features improves the performance of an AS model compared to a baseline model using established linguistic features.

The study's findings contribute empirical evidence regarding the practical utility of LB frequency as a feature in AS, informing efforts to enhance the accuracy and construct representation of automated writing evaluation tools.

\section{Background and Related Work}
\label{sec:background}

\subsection{Automated Scoring of Writing}
AS systems typically operate by extracting a range of linguistic features from essays and using machine learning algorithms to build a model that predicts scores based on these features, trained on human-scored examples \citep{Shermis2013}. Feature engineering is critical; early systems focused on surface features, while modern systems like e-rater incorporate measures related to grammar, usage, mechanics, style, organization, development, lexical complexity, and syntactic variety \citep{Attali2006, Chen2016}. The goal is to select features that not only correlate with scores but also represent theoretically meaningful aspects of the writing construct. Evaluating AS models often involves measuring agreement with human raters using metrics like exact agreement, Cohen's Kappa, and Quadratic Weighted Kappa (QWK), with QWK > 0.70 often considered a benchmark for operational systems \citep{Ramineni2013}.

\subsection{Lexical Bundles in L1 and L2 Writing}
LBs are identified computationally based on frequency and distribution criteria within a corpus \citep{Biber1999}. They differ from idioms or collocations in that they are not necessarily semantically opaque or grammatically fixed units but rather reflect conventionalized ways of expressing meaning and structuring discourse. Research has shown LBs are characteristic of specific registers, like academic prose and conversation \citep{Biber2007, Biber2004}. Studies comparing L1 and L2 writing have found differences in LB usage patterns \citep{Chen2016b}. Within L2 writing research, studies have linked LB frequency and type to proficiency. For instance, \citet{Staples2013} found higher proficiency TOEFL writers used fewer 4-word LBs and fewer prompt-related LBs. \citet{Appel2016} and \citet{Vo2019} similarly found inverse relationships between overall LB frequency and proficiency scores in different assessment contexts, suggesting less proficient writers rely more heavily on these formulaic sequences. This body of work establishes LBs as a relevant linguistic construct related to L2 writing development and proficiency.

\section{Method}
\label{sec:method}

\subsection{Corpus and Sampling}
This study utilized the TOEFL11 corpus \citep{Blanchard2014}, a collection of 12,100 essays written for the TOEFL independent writing task between 2006-2007 by test-takers from 11 distinct L1 backgrounds (Arabic, Chinese, French, German, Hindi, Italian, Japanese, Korean, Spanish, Telugu, Turkish). Essays were scored by ETS-trained raters on a 0-5 scale and subsequently categorized into Low, Medium, and High proficiency levels for the corpus release. Eight different writing prompts were used across the dataset.

To ensure balanced comparisons across L1 backgrounds and proficiency levels, and mitigate potential bias from skewed distributions, essays from German and Hindi speakers were excluded (due to unusually low counts in the Low proficiency category). Essays shorter than nine words were also removed as they were too short for meaningful 9-gram analysis. Following these exclusions, a stratified random sampling procedure was implemented. Aiming to match word counts across the remaining nine L1s and three score levels, the stratum with the lowest total word count (French-Low) was used as the target. This resulted in a final sampled subcorpus of 1,225 essays, totaling 335,687 words. Details of the sampled corpus composition are provided in Table \ref{tab:corpus_sampled}.

% --- Table 4 ---
\begin{table}[htbp]
  \centering
  \caption{Sampled Corpus Essay Word Counts and Number of Essays per L1 per Score Level}
  \label{tab:corpus_sampled}
  \begin{tabular}{lcccccc}
    \toprule
    L1  & Low        & Low       & Medium     & Medium    & High       & High      \\
        & (\# Words) & (\# Essays) & (\# Words) & (\# Essays) & (\# Words) & (\# Essays) \\
    \midrule
    ARA & 12,478     & 61        & 12,454     & 41        & 12,297     & 35        \\
    FRA & 12,479     & 63        & 12,477     & 41        & 12,471     & 36        \\
    ITA & 12,472     & 63        & 12,442     & 41        & 12,472     & 38        \\
    JPN & 12,471     & 72        & 12,388     & 44        & 12,282     & 33        \\
    KOR & 12,461     & 66        & 12,426     & 44        & 12,445     & 34        \\
    SPA & 12,465     & 55        & 12,456     & 41        & 12,399     & 36        \\
    TEL & 12,435     & 44        & 12,404     & 38        & 12,419     & 32        \\
    TUR & 12,478     & 62        & 12,477     & 40        & 12,378     & 34        \\
    ZHO & 12,479     & 57        & 12,458     & 39        & 12,324     & 33        \\
    \midrule
    Total & 112,218   & 543       & 111,982    & 371       & 111,487    & 311       \\
    \bottomrule
  \end{tabular}
\end{table}
% -------------

\subsection{Lexical Bundle Extraction and Features}
LBs were extracted from the sampled corpus using the Lexical Bundle Identification and Analysis Program (LBiaP; \citet{Cortes2023}). This Python-based tool identifies n-grams (sequences of n words) ranging from 3 to 9 words in length that meet predefined frequency and range (document distribution) thresholds, based on established corpus linguistic methods \citep{Biber1999, Cortes2004}. The specific criteria used are detailed in Table \ref{tab:lb_criteria}. A key feature of LBiaP is its method for handling potential overlap between bundles during the counting process, ensuring more accurate frequency measures.

% --- Table 6 ---
\begin{table}[htbp]
  \centering
  \caption{LBiaP Frequency and Range Criteria for LBs of Each Length}
  \label{tab:lb_criteria}
  \begin{tabular}{lcc}
    \toprule
    LB category & Frequency (per million words) & Range (min \# files) \\
    \midrule
    9-word      & 5                             & 5                    \\
    8-word      & 5                             & 5                    \\
    7-word      & 5                             & 5                    \\
    6-word      & 5                             & 5                    \\
    5-word      & 10                            & 5                    \\
    4-word      & 20                            & 5                    \\
    3-word      & 40                            & 5                    \\
    \bottomrule
  \end{tabular}
\end{table}
% -------------

Following extraction, LBs were classified into two types based on their occurrence relative to the eight different writing prompts in the corpus:
\begin{itemize}
    \item \textbf{Prompt LBs:} Bundles found exclusively in essays written in response to one specific prompt.
    \item \textbf{Non-prompt LBs:} Bundles found in essays corresponding to two or more different prompts, suggesting more general usability.
\end{itemize}
This categorization yielded 132 non-prompt and 151 prompt LBs, totaling 283 unique bundles. For each essay, frequencies for prompt, non-prompt, and total LBs (both overall and per specific length) were calculated and subsequently normalized per 100 words to account for variations in essay length. Initial checks using the Shapiro-Wilk test indicated non-normal distributions for these frequency counts. Therefore, Kruskal-Wallis and Mann-Whitney U tests (with a Bonferroni correction setting $\alpha$ = .017 for pairwise comparisons) were employed to investigate frequency differences across the three proficiency score levels, addressing the first research question.

\subsection{Baseline Linguistic Features}
A set of baseline linguistic features, intended to represent common categories included in AS systems like e-rater \citep{Chen2016}, was extracted using publicly available, validated NLP tools:
\begin{itemize}
    \item \textbf{GAMET \citep{Crossley2019b}:} Used to calculate frequencies of mechanical (e.g., spelling, capitalization, punctuation) and grammatical errors.
    \item \textbf{TAACO \citep{Crossley2016, Crossley2019}:} Used to measure lexical and semantic overlap between adjacent sentences and paragraphs, providing indices of local text cohesion.
    \item \textbf{TAASSC \citep{Kyle2016}:} Used to compute measures of syntactic complexity (e.g., variability in clause structures) and lexical sophistication (e.g., use of academic vocabulary, word frequency).
\end{itemize}
Notably, as TAACO and TAASSC rely on standard punctuation for sentence parsing, a preparatory script was developed and applied to temporarily correct non-standard sentence-final punctuation observed in some essays to enable feature extraction; this practical necessity carries implications discussed later. From the features generated by these tools, a subset was selected for the baseline model based on their moderate correlation with the human-assigned proficiency scores (see Table \ref{tab:baseline_features}).

% --- Table 8 ---
\begin{table}[htbp]
  \centering
  \caption{Selected Baseline Essay Features Analyzed for Automated Essay Scoring}
  \label{tab:baseline_features}
  \small % Reduce font size slightly for table
  \begin{tabular}{lcl}
    \toprule
    Feature name                     & Correlation with score & Corresponding e-rater feature category \\
    \midrule
    misspelling\_per\_100\_words      & -0.513                 & Usage                                  \\
    adjacent\_overlap\_all\_para\_div\_seg & 0.461                  & Development                            \\
    lexical\_density\_type           & 0.380                  & Positive features                      \\
    grammar\_per\_100\_words          & -0.356                 & Grammar                                \\
    typographical\_per\_100\_words    & -0.290                 & Mechanics                              \\
    acad\_collexeme\_ratio\_type      & 0.231                  & Positive features                      \\
    adjacent\_overlap\_binary\_all\_sent & 0.228                  & Development                            \\
    nsubj\_stdev                     & 0.209                  & Development                            \\
    pobj\_stdev                      & 0.203                  & Development                            \\
    dobj\_stdev                      & 0.193                  & Development                            \\
    all\_connective                  & 0.102                  & Organization                           \\
    \bottomrule
  \end{tabular}
\end{table}
% -------------

\subsection{Data Analysis}
\label{sec:analysis}
The data analysis proceeded in two stages corresponding to the research questions.

\textbf{RQ1 Analysis:} To investigate the relationship between LB frequency and proficiency level, statistical comparisons were performed on the normalized LB frequency counts. As Shapiro-Wilk tests indicated non-normality, non-parametric tests were employed. The Kruskal-Wallis H test was used to determine if there were statistically significant differences in the median frequency of each LB category (e.g., 3-word non-prompt, total prompt, overall total) across the three proficiency groups (Low, Medium, High). Where the Kruskal-Wallis test indicated a significant overall difference ($p < .05$), post-hoc Mann-Whitney U tests were conducted for pairwise comparisons between proficiency levels (Low vs. Medium, Low vs. High, Medium vs. High). A Bonferroni correction was applied to the alpha level for these pairwise tests (adjusted $\alpha = 0.05 / 3 = 0.017$) to control for multiple comparisons. Effect sizes ($\eta^2$ based on H statistic) were also calculated for the Kruskal-Wallis tests.

\textbf{RQ2 Analysis:} To evaluate the contribution of LB features to the AS model, Support Vector Machine (SVM) classifiers were trained and tested. SVMs are supervised learning algorithms effective for classification tasks, seeking an optimal hyperplane to separate data points belonging to different classes \citep{Cortes1995}; they have shown strong performance in previous e-rater modeling studies \citep{Chen2016}. Two SVM models (using a linear kernel) were compared: a Baseline Model using only the features in Table \ref{tab:baseline_features}, and an Extended Model using the baseline features plus the three aggregate normalized LB frequency counts (total prompt, total non-prompt, overall total). This strategy of using aggregate totals, rather than selecting specific LB lengths or types based on RQ1 results, was chosen deliberately to capture the overall contribution of LB frequency across categories without introducing potential subjectivity in feature selection based on preliminary analyses, aiming instead for a broad assessment of the combined impact of LB frequency. Model performance was evaluated using stratified 3-fold cross-validation, ensuring that each fold maintained the original distribution of L1 backgrounds and proficiency levels. The models' predictions (Low, Medium, High) were compared against the actual human-assigned labels using:
\begin{itemize}
    \item \textbf{Exact Agreement:} The proportion of essays correctly classified.
    \item \textbf{Cohen’s Kappa (K):} Agreement corrected for chance.
    \item \textbf{Quadratic Weighted Cohen’s Kappa (QWK):} Agreement corrected for chance, assigning partial credit based on the severity of misclassification (closer predictions receive higher scores).
\end{itemize}

\section{Results}
\label{sec:results}

\subsection{RQ1: Relationship Between LB Frequency and Score}
The statistical analyses addressing the first research question revealed significant differences in the usage frequency of several LB categories across the Low, Medium, and High proficiency groups (see Table \ref{tab:lb_results_landscape} for full results). Kruskal-Wallis tests showed significant overall differences ($p < .05$) for 3-word LBs (non-prompt, prompt, and total), 4-word LBs (prompt and total), 6-word non-prompt LBs, and 8-word prompt LBs. The aggregate frequency measures for total non-prompt LBs and overall total LBs also differed significantly across groups. Post-hoc Mann-Whitney U tests indicated that these differences often occurred between the Low and High groups, or Low and Medium groups. Notably, the aggregate non-prompt and overall total LB frequencies showed significant differences between the Medium and High groups as well. However, frequencies for 5-word, 7-word, and 9-word LBs generally did not show statistically significant variation across proficiency levels. While statistically significant, the calculated effect sizes for these differences were generally small ($\eta^2$ mostly $\le 0.02$).

Observing the mean frequencies (Table \ref{tab:lb_results_landscape}) suggested qualitative trends: lower-scoring essays tended to have slightly higher overall LB frequencies. Higher-scoring essays appeared to use relatively more 3- and 4-word prompt LBs, whereas lower-scoring essays showed comparatively higher usage of longer prompt LBs (5-9 words).

% --- Table 4 (Landscape) ---
% Make sure \usepackage{rotating} is in your preamble
\begin{sidewaystable}[htbp] % Use sidewaystable environment
  \centering
  \caption{Analysis Results of Various LB Features Based on Score Level}
  \label{tab:lb_results_landscape} % Use a unique label if needed
  \tiny % Keep font size small, adjust if possible after viewing landscape
  \begin{tabular}{lccccccccc}
    \toprule
     & \multicolumn{3}{c}{\textbf{Kruskal Wallis}} & \multicolumn{3}{c}{\textbf{Feature Means}} & \multicolumn{3}{c}{\textbf{Mann-Whitney U}} \\
    \cmidrule(lr){2-4} \cmidrule(lr){5-7} \cmidrule(lr){8-10} % Lines under multi-column headers
    Feature & H statistic & p-value & effect size & Low & Medium & High & Low vs High p & Low vs Med p & Med vs High p \\
    \midrule
    \textbf{3-word non-prompt bundles} & 9.867 & 0.007* & 0.008 & 0.0167 & 0.0176 & 0.0153 & 0.426 & 0.021 & 0.002** \\
    \textbf{3-word prompt bundles}     & 11.692 & 0.003* & 0.010 & 0.0020 & 0.0020 & 0.0022 & 0.001** & 0.124 & 0.086 \\
    \textbf{3-word total}              & 8.069 & 0.018* & 0.007 & 0.0187 & 0.0196 & 0.0175 & 0.801 & 0.017** & 0.009** \\
    \midrule
    \textbf{4-word non-prompt bundles} & 6.932 & 0.031* & 0.006 & 0.0031 & 0.0030 & 0.0028 & 0.037 & 0.022 & 0.471 \\
    \textbf{4-word prompt bundles}     & 25.810 & 0.000* & 0.021 & 0.0008 & 0.0013 & 0.0014 & 0.000** & 0.003** & 0.057 \\
    \textbf{4-word total}              & 13.477 & 0.001* & 0.011 & 0.0039 & 0.0043 & 0.0042 & 0.001** & 0.005** & 0.978 \\
    \midrule
    \textbf{5-word non-prompt bundles} & 0.112 & 0.946 & 0.000 & 0.0006 & 0.0004 & 0.0003 & 0.757 & 0.959 & 0.780 \\
    \textbf{5-word prompt bundles}     & 0.148 & 0.929 & 0.000 & 0.0010 & 0.0007 & 0.0007 & 0.756 & 0.752 & 0.863 \\
    \textbf{5-word total}              & 0.499 & 0.779 & 0.000 & 0.0016 & 0.0012 & 0.0010 & 0.540 & 0.793 & 0.547 \\
    \midrule
    \textbf{6-word non-prompt bundles} & 14.951 & 0.001* & 0.012 & 0.0001 & 0.0002 & 0.0003 & 0.000** & 0.004** & 0.397 \\
    \textbf{6-word prompt bundles}     & 1.203 & 0.548 & 0.001 & 0.0012 & 0.0010 & 0.0007 & 0.853 & 0.336 & 0.349 \\
    \textbf{6-word total}              & 2.470 & 0.291 & 0.002 & 0.0013 & 0.0012 & 0.0010 & 0.331 & 0.142 & 0.473 \\
    \midrule
    \textbf{7-word non-prompt bundles} & N/A & N/A & N/A & 0.0000 & 0.0000 & 0.0000 & N/A & N/A & N/A \\
    \textbf{7-word prompt bundles}     & 0.419 & 0.811 & 0.000 & 0.0006 & 0.0005 & 0.0004 & 0.726 & 0.539 & 0.743 \\
    \textbf{7-word total}              & 0.419 & 0.811 & 0.000 & 0.0006 & 0.0005 & 0.0004 & 0.726 & 0.539 & 0.743 \\
    \midrule
    \textbf{8-word non-prompt bundles} & N/A & N/A & N/A & 0.0000 & 0.0000 & 0.0000 & N/A & N/A & N/A \\
    \textbf{8-word prompt bundles}     & 8.224 & 0.016* & 0.007 & 0.0005 & 0.0003 & 0.0004 & 0.004** & 0.171 & 0.179 \\
    \textbf{8-word total (same as prompt)} & 8.224 & 0.016* & 0.007 & 0.0005 & 0.0003 & 0.0004 & 0.004** & 0.171 & 0.179 \\
    \midrule
    \textbf{9-word non-prompt bundles} & N/A & N/A & N/A & 0.0000 & 0.0000 & 0.0000 & N/A & N/A & N/A \\
    \textbf{9-word prompt bundles}     & 3.283 & 0.194 & 0.003 & 0.0027 & 0.0015 & 0.0013 & 0.144 & 0.135 & 0.523 \\
    \textbf{9-word total (same as prompt)} & 3.283 & 0.194 & 0.003 & 0.0027 & 0.0015 & 0.0013 & 0.144 & 0.135 & 0.523 \\
    \midrule
    \textbf{Total (prompt)}            & 0.763 & 0.683 & 0.001 & 0.0088 & 0.0073 & 0.0071 & 0.464 & 0.483 & 0.815 \\
    \textbf{Total (non-prompt)}        & 9.365 & 0.009* & 0.008 & 0.0205 & 0.0213 & 0.0186 & 0.318 & 0.033 & 0.002** \\
    \textbf{Total (overall)}           & 8.618 & 0.013* & 0.007 & 0.0293 & 0.0285 & 0.0258 & 0.025 & 0.515 & 0.003** \\
    \midrule
    \multicolumn{10}{l}{\footnotesize{*$p < .05$. **$p < .017$ (Bonferroni corrected for Mann-Whitney U pairwise comparisons)}} \\ % Clarified footnote
    \bottomrule
  \end{tabular}
\end{sidewaystable} % End sidewaystable environment
% -------------

\subsection{RQ2: Impact of LB Features on AS Model Performance}
The comparison of the SVM models demonstrated clear benefits from including the aggregate LB frequency features.

\textbf{Overall Model Performance (Table \ref{tab:overall_perf}):} The Extended Model achieved higher scores on all overall evaluation metrics compared to the Baseline Model. Exact agreement improved from 0.677 to 0.696 (+2.81\%). Cohen’s Kappa increased from 0.515 to 0.544 (+5.63\%), signifying better-than-chance agreement. Quadratic Weighted Kappa (QWK) improved from 0.731 to 0.746 (+2.05\%). Both models met the common QWK > 0.70 benchmark \citep{Ramineni2013}, but the addition of LB features resulted in stronger alignment with human scores.

% --- Table 10 ---
\begin{table}[htbp]
  \centering
  \caption{Overall Comparison of Baseline and Extended Models}
  \label{tab:overall_perf}
  \begin{tabular}{lccc}
    \toprule
    Metric               & Baseline Model & Extended Model & Change (\%) \\
    \midrule
    Exact Agreement      & 0.677          & 0.696          & 2.81        \\
    Cohen’s K            & 0.515          & 0.544          & 5.63        \\
    Quadratic Cohen’s K & 0.731          & 0.746          & 2.05        \\
    \bottomrule
  \end{tabular}
\end{table}
% -------------

\textbf{Score-Specific Model Performance (Table \ref{tab:score_perf}):} Analyzing performance for each proficiency level revealed that the improvement was not uniform. The most substantial gains were realized for essays in the Low and Medium proficiency categories. For Low essays, exact agreement increased from 0.761 to 0.838 (+10.12\%), and Kappa improved from 0.608 to 0.640 (+5.26\%). For Medium essays, while exact agreement rose modestly (+3.04\%), Kappa showed a large relative increase (+14.34\%) from 0.286 to 0.327. In contrast, performance for High proficiency essays saw minimal change, with slight increases in exact agreement (+0.60\%) and Kappa (+1.88\%).

% --- Table 11 ---
\begin{table}[htbp]
  \centering
  \caption{Score-Specific Comparison of Baseline and Extended Model}
  \label{tab:score_perf}
  \begin{tabular}{llccc}
    \toprule
    Proficiency Level & Metric          & Baseline Model & Extended Model & Change (\%) \\
    \midrule
    Low               & Exact Agreement & 0.761          & 0.838          & 10.12       \\
    Low               & Cohen’s K       & 0.608          & 0.640          & 5.26        \\
    Medium            & Exact Agreement & 0.691          & 0.712          & 3.04        \\
    Medium            & Cohen’s K       & 0.286          & 0.327          & 14.34       \\
    High              & Exact Agreement & 0.837          & 0.842          & 0.60        \\
    High              & Cohen’s K       & 0.640          & 0.652          & 1.88        \\
    \bottomrule
  \end{tabular}
\end{table}
% -------------

\section{Discussion}
\label{sec:discussion}
This study provides initial empirical evidence supporting the integration of aggregate lexical bundle frequency features into automated scoring models for L2 writing, moving beyond prior suggestions in the literature. The study's findings indicate a statistically significant, albeit complex, relationship between the frequency of specific LB categories and scored proficiency levels (RQ1). However, it is important to acknowledge that the effect sizes associated with these significant differences were generally small (Table \ref{tab:lb_results_landscape}). This likely reflects a confluence of factors: the study's substantial statistical power allowing detection of subtle differences; the inherent nature of LBs representing only one facet of multifaceted writing proficiency where qualitative aspects like appropriate use may be as important as raw frequency; potential homogenizing effects of specific writing prompts; and the methodological consequence of using broad, collapsed score categories which can obscure finer distinctions and increase within-group variance. Furthermore, the interpretation of 'non-prompt' status, defined here as appearing across two or more of the eight prompts, warrants caution. Overlap could arise from shared thematic elements between prompts (e.g., prompts P2 and P3 in the TOEFL11 corpus both concern 'young people') or common academic argumentative structures, rather than solely reflecting transferable lexical resources independent of specific prompt content. A deeper functional analysis would be needed to fully disentangle these possibilities. While the observed trend for lower-scoring essays to utilize more LBs overall, particularly longer prompt-related ones, offers tentative support for hypotheses regarding formulaic language reliance \citep{Ellis2002, Staples2013}, this relationship warrants deeper investigation beyond simple frequency counts.

Despite the small individual effect sizes for specific LB categories, the inclusion of the three aggregate LB frequency features yielded notable improvements in AS model performance (RQ2), enhancing overall agreement with human raters, particularly when assessed by QWK (Table \ref{tab:overall_perf}). This key finding suggests that while individual LB categories may have limited discriminatory power on their own, their combined pattern across prompt, non-prompt, and overall usage provides unique predictive information not fully captured by the baseline linguistic features, thereby contributing to the improved multivariate model performance. The score-specific analysis (Table \ref{tab:score_perf}) was particularly revealing, demonstrating that the performance gains were primarily driven by enhanced classification of Low and Medium proficiency essays. This differential impact suggests that aggregate LB frequency features may be especially valuable for distinguishing characteristics of developing L2 writing, perhaps related to fluency or reliance on formulaic structures, which might be less differentiating or already captured by other linguistic markers in high-proficiency writing.

While the magnitude of the QWK improvement (+2.05\%) might appear modest, its practical significance in a large-scale, high-stakes testing context like TOEFL should be considered. Processing hundreds of thousands of essays requires significant resources for human rating, especially when resolving disagreements between an initial human rater and the AS system. Even a small percentage increase in human-machine agreement, as indicated by the improved QWK, could translate into substantial savings in time and cost by reducing the number of essays requiring expensive second human ratings. These results therefore affirm the potential value of integrating aggregate LB frequency analysis into AS systems like e-rater. Nonetheless, it is crucial to contextualize these findings relative to the baseline model used. Operational AS systems undergo continuous development and utilize proprietary feature sets that are likely more extensive than the baseline model employed in this research \citep{Attali2006, Quinlan2009}. Consequently, the incremental predictive value demonstrated here requires further validation against more complex, potentially operational, baseline models.

\section{Limitations and Future Directions}
\label{sec:limitations}
Several limitations should be considered when interpreting these findings. The baseline model, while aligned with published descriptions, is less comprehensive than proprietary, operational AS systems, and feature selection relied primarily on correlation, a relatively basic approach. The use of collapsed score categories limits fine-grained analysis and likely contributed to the small effect sizes observed. The LB extraction relied on frequency and range criteria established in prior corpus research (\ref{tab:lb_criteria}); while conventional and implemented directly via the LBiaP tool to avoid subjective adjustments for this study, these thresholds might not be perfectly optimized for this specific dataset's size (~335k words vs. per-million norms). Different criteria, potentially derived empirically or adaptively set based on corpus size (an option for future work), could yield different LB sets and results. Furthermore, this study focused exclusively on LB frequency. A further methodological limitation involved the necessity of a pre-processing script to correct non-standard punctuation before applying certain feature extraction tools (TAACO, TAASSC). This workaround highlights a practical challenge for real-time AS, as punctuation errors were particularly prevalent in lower- and medium-scoring essays (supported also by the negative correlation of the typographical error feature) and relying on such corrections might mask genuine proficiency issues related to mechanics. Finally, the sample size, although adequate for this report's analyses, might limit the generalizability of specific SVM model parameters compared to larger operational development datasets.

Future research should aim to incorporate functional/structural LB analyses, use the original score scale if available, and explore L1 interactions. Testing LB features with various machine learning algorithms would clarify if gains are algorithm-dependent. Evaluating LBs within richer feature sets will better determine their unique contribution. Ultimately, these findings suggest AS developers should consider experimenting with LB frequency, particularly noting its potential to better differentiate developing writers, as a valuable candidate feature for inclusion in future system refinements, weighing the demonstrated performance gains against the computational considerations of implementation.

\section{Conclusion}
\label{sec:conclusion}
This study investigated the role of lexical bundle frequency in relation to L2 writing proficiency and its utility as a feature in automated scoring. Significant relationships were found between the frequency of specific LB types and human-assigned proficiency scores for TOEFL essays, although the magnitude of these differences (effect sizes) was generally small. Critically, integrating aggregate LB frequency features (total prompt, total non-prompt, overall total) into an SVM-based AS model improved agreement with human raters, particularly for identifying essays at Low and Medium proficiency levels. These findings provide initial empirical support for further experimentation incorporating measures of phraseological frequency into AS systems to potentially enhance their accuracy and linguistic grounding, offering a valuable direction for future AS development and research.

% --- Optional Sections ---
% \section*{Acknowledgements}
% (Add acknowledgements here if needed)

\section*{Data Availability Statement}
The TOEFL11 corpus used in this study is available from the Linguistic Data Consortium (LDC) catalog number LDC2014T06.

\section*{Code Availability Statement}
Code for LB extraction using LBiaP is available at \url{https://github.com/lbiap/main}.

% --- Bibliography ---
\bibliography{references} % Link to your .bib file

\begin{thebibliography}{22}
\providecommand{\natexlab}[1]{#1}
\providecommand{\url}[1]{\texttt{#1}}
\expandafter\ifx\csname urlstyle\endcsname\relax
  \providecommand{\doi}[1]{doi: #1}\else
  \providecommand{\doi}{doi: \begingroup \urlstyle{rm}\Url}\fi

\bibitem[Appel and Wood(2016)]{Appel2016}
Randy Appel and David Wood.
\newblock Recurrent word combinations in eap test-taker writing: Differences between high-and low-proficiency levels.
\newblock \emph{Language Assessment Quarterly}, 13\penalty0 (1):\penalty0 55--71, 2016.

\bibitem[Attali and Burstein(2006)]{Attali2006}
Yigal Attali and Jill Burstein.
\newblock Automated essay scoring with e-rater v. 2.
\newblock \emph{The Journal of Technology, Learning and Assessment}, 4\penalty0 (3), 2006.

\bibitem[Biber and Barbieri(2007)]{Biber2007}
Douglas Biber and Federica Barbieri.
\newblock Lexical bundles in university spoken and written registers.
\newblock \emph{English for specific purposes}, 26\penalty0 (3):\penalty0 263--286, 2007.

\bibitem[Biber et~al.(1999)Biber, Johansson, Leech, Conrad, and Finegan]{Biber1999}
Douglas Biber, Stig Johansson, Geoffrey Leech, Susan Conrad, and Edward Finegan.
\newblock \emph{Longman grammar of spoken and written English}.
\newblock Pearson Education Limited, 1999.

\bibitem[Biber et~al.(2004)Biber, Conrad, and Cortes]{Biber2004}
Douglas Biber, Susan Conrad, and Viviana Cortes.
\newblock If you look at…: Lexical bundles in university teaching and textbooks.
\newblock \emph{Applied linguistics}, 25\penalty0 (3):\penalty0 371--405, 2004.

\bibitem[Blanchard et~al.(2014)Blanchard, Tetreault, Higgins, Cahill, and Chodorow]{Blanchard2014}
Daniel Blanchard, Joel Tetreault, Derrick Higgins, Aoife Cahill, and Martin Chodorow.
\newblock {ETS Corpus of Non-Native Written English}.
\newblock Linguistic Data Consortium, 2014.
\newblock URL \url{https://catalog.ldc.upenn.edu/LDC2014T06}.

\bibitem[Chen et~al.(2016)Chen, Fife, Bejar, and Rupp]{Chen2016}
Jing Chen, James~H Fife, Isaac~I Bejar, and Andr{\'e}~A Rupp.
\newblock Building e-rater® scoring models using machine learning methods.
\newblock \emph{ETS Research Report Series}, 2016\penalty0 (1):\penalty0 1--12, 2016.

\bibitem[Chen and Baker(2016)]{Chen2016b}
Yu-Hua Chen and Paul Baker.
\newblock Investigating criterial discourse features across second language development: Lexical bundles in rated learner essays, cefr b1, b2 and c1.
\newblock \emph{Applied Linguistics}, 37\penalty0 (6):\penalty0 849--880, 2016.

\bibitem[Cortes and Vapnik(1995)]{Cortes1995}
Corinna Cortes and Vladimir Vapnik.
\newblock Support-vector networks.
\newblock \emph{Machine learning}, 20\penalty0 (3):\penalty0 273--297, 1995.

\bibitem[Cortes(2004)]{Cortes2004}
Viviana Cortes.
\newblock Lexical bundles in published and student disciplinary writing: Examples from history and biology.
\newblock \emph{English for specific purposes}, 23\penalty0 (4):\penalty0 397--423, 2004.

\bibitem[Cortes and Lake(2023)]{Cortes2023}
Viviana Cortes and William~B Lake.
\newblock Lbiap: A solution to the problem of attaining observation independence in lexical bundle studies.
\newblock \emph{International Journal of Corpus Linguistics}, 28\penalty0 (2):\penalty0 263--277, 2023.

\bibitem[Crossley et~al.(2016)Crossley, Kyle, and McNamara]{Crossley2016}
Scott~A Crossley, Kristopher Kyle, and Danielle~S McNamara.
\newblock The tool for the automatic analysis of text cohesion (taaco): Automatic assessment of local, global, and text cohesion.
\newblock \emph{Behavior research methods}, 48\penalty0 (4):\penalty0 1227--1237, 2016.

\bibitem[Crossley et~al.(2019{\natexlab{a}})Crossley, Bradfield, and Bustamante]{Crossley2019b}
Scott~A Crossley, Franklin Bradfield, and Analynn Bustamante.
\newblock Using human judgments to examine the validity of automated grammar, syntax, and mechanical errors in writing.
\newblock \emph{Journal of Writing Research}, 11\penalty0 (2), 2019{\natexlab{a}}.

\bibitem[Crossley et~al.(2019{\natexlab{b}})Crossley, Kyle, and Dascalu]{Crossley2019}
Scott~A Crossley, Kristopher Kyle, and Mihai Dascalu.
\newblock The tool for the automatic analysis of cohesion 2.0: Integrating semantic similarity and text overlap.
\newblock \emph{Behavior research methods}, 51\penalty0 (1):\penalty0 14--27, 2019{\natexlab{b}}.

\bibitem[Ellis(2002)]{Ellis2002}
Nick~C Ellis.
\newblock Frequency effects in language processing: A review with implications for theories of implicit and explicit language acquisition.
\newblock \emph{Studies in second language acquisition}, 24\penalty0 (2):\penalty0 143--188, 2002.

\bibitem[Kyle(2016)]{Kyle2016}
Kristopher Kyle.
\newblock \emph{Measuring syntactic development in L2 writing: Fine grained indices of syntactic complexity and usage-based indices of syntactic sophistication}.
\newblock PhD thesis, Georgia State University, 2016.

\bibitem[Quinlan et~al.(2009)Quinlan, Higgins, and Wolff]{Quinlan2009}
Thomas Quinlan, Derrick Higgins, and Susan Wolff.
\newblock Evaluating the construct-coverage of the e-rater® scoring engine.
\newblock Technical Report RR-09-01, ETS Research Report Series, 2009.
\newblock URL \url{https://onlinelibrary.wiley.com/doi/abs/10.1002/j.2333-8504.2009.tb02158.x}.

\bibitem[Ramineni and Williamson(2013)]{Ramineni2013}
Chaitanya Ramineni and David~M Williamson.
\newblock Automated essay scoring: Psychometric guidelines and practices.
\newblock \emph{Assessing Writing}, 18\penalty0 (1):\penalty0 25--39, 2013.

\bibitem[Shermis and Burstein(2013)]{Shermis2013}
Mark~D Shermis and Jill Burstein, editors.
\newblock \emph{Handbook of automated essay evaluation: Current applications and new directions}.
\newblock Routledge, 2013.

\bibitem[Staples et~al.(2013)Staples, Egbert, Biber, and McClair]{Staples2013}
Shelley Staples, Jesse Egbert, Douglas Biber, and Alyson McClair.
\newblock Formulaic sequences and eap writing development: Lexical bundles in the toefl ibt writing section.
\newblock \emph{Journal of English for Academic Purposes}, 12\penalty0 (3):\penalty0 214--225, 2013.

\bibitem[Vo(2019)]{Vo2019}
Sonca Vo.
\newblock Use of lexical features in non-native academic writing.
\newblock \emph{Journal of Second Language Writing}, 44:\penalty0 1--12, 2019.

\bibitem[Xi(2010)]{Xi2010}
Xiaoming Xi.
\newblock Automated scoring and feedback systems: Where are we and where are we heading?
\newblock \emph{Language testing}, 27\penalty0 (3):\penalty0 291--300, 2010.

\end{thebibliography}

\end{document}